\documentclass[letterpaper, 10 pt, conference]{ieeeconf}  

\usepackage{url}
\usepackage{graphicx}
\usepackage{booktabs}
\usepackage{romannum}
\usepackage{amsfonts}
\usepackage{multirow}

\usepackage{array}
\usepackage{makecell}
\usepackage{float}
\usepackage[export]{adjustbox} 
\usepackage{algorithm}
\usepackage{balance}
\usepackage{cite}
\usepackage{algpseudocode} 
\usepackage{longtable}
\usepackage{threeparttable}

\usepackage{booktabs}

\usepackage{graphics}
\usepackage{upgreek}
\usepackage{amssymb}
\usepackage{refstyle}

\usepackage{bbding}

\usepackage{amsmath}
\usepackage[table,dvipsnames]{xcolor}

\usepackage{subcaption}
\usepackage{mathtools}

\definecolor{c1_red}{HTML}{dc3023}
\definecolor{c1_green}{HTML}{0c8918}

\usepackage[breaklinks,colorlinks]{hyperref}

\usepackage{cleveref}

\IEEEoverridecommandlockouts                              

\title{\LARGE \bf
\makebox[0pt][l]{%
    \smash{\raisebox{-0.5\height}{\includegraphics[height=1.2cm]{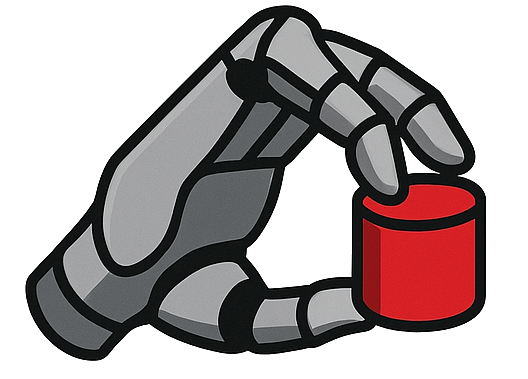}}%
}}%
\hspace{1.5cm} %
DORA: \textcolor{red}{O}bject \textcolor{red}{A}ffordance-Guided \textcolor{red}{R}einforcement Learning \\for \textcolor{red}{D}exterous Robotic Manipulation
}
\author{Lei Zhang$^{1,2\dag}$, Soumya Mondal$^{2,3}$, Zhenshan Bing$^{3\dag}$, Kaixin Bai$^{1,2}$, Diwen Zheng$^{2,3}$, \\Zhaopeng Chen$^{2}$, Alois Christian Knoll$^{3}$, Jianwei Zhang$^{1}$
\thanks{\dag Corresponding author. lei.zhang-1@studium.uni-hamburg.de, zhenshan.bing@tum.de}
\thanks{{$^{1}$TAMS (Technical Aspects of Multimodal Systems), Department of
Informatics, Universit\"at Hamburg, Hamburg, Germany}, {$^{2}$Agile Robots AG, Munich, Germany}, {$^{3}$Technical University of Munich, Munich, Germany}.}
}
\DeclarePairedDelimiterX{\norm}[1]{\lVert}{\rVert}{#1}

\begin{document}

\maketitle

\thispagestyle{empty}
\pagestyle{empty}

\begin{abstract}    

Dexterous robotic manipulation remains a longstanding challenge in robotics due to the high dimensionality of control spaces and the semantic complexity of object interaction. In this paper, we propose an object affordance-guided reinforcement learning (RL) framework that enables a multi-fingered robotic hand to learn human-like manipulation strategies more efficiently. By leveraging object affordance maps, our approach generates semantically meaningful grasp pose candidates that serve as both policy constraints and priors during training. We introduce a voting-based grasp classification mechanism to ensure functional alignment between grasp configurations and object affordance regions. Furthermore, we incorporate these constraints into a generalizable RL pipeline and design a reward function that unifies affordance-awareness with task-specific objectives. Experimental results across three manipulation tasks—cube grasping, jug grasping and lifting, and hammer use—demonstrate that our affordance-guided approach improves task success rates by an average of 15.4\% compared to baselines. These findings highlight the critical role of object affordance priors in enhancing sample efficiency and learning generalizable, semantically grounded manipulation policies. For more details, please visit our project website: \href{https://sites.google.com/view/dora-manip}{https://sites.google.com/view/dora-manip}.

\end{abstract}

\section{Introduction}
\label{intro}

Dexterous manipulation is a fundamental yet challenging capability for robotic systems aiming to interact with real-world objects in a human-like manner~\cite{andrychowicz2020learning}. Unlike rigid industrial grippers, multi-fingered robotic hands possess a high number of degrees of freedom (DoFs), offering the potential for flexible and nuanced manipulation. However, learning to control such systems remains difficult due to the complexity of hand-object interactions, high-dimensional control spaces, and the need for semantic understanding of object parts~\cite{mandikal2021learning,zhang2024multi,wu2023learning}.

Human manipulation behavior is inherently guided by object affordances—the action possibilities provided by specific regions of an object~\cite{gibson1979ecological, do2018affordancenet}. For instance, when pouring water from a pitcher, humans intuitively grasp the handle due to its functional role. Inspired by this observation, we propose a RL framework that integrates object affordance knowledge into both grasp planning and policy training for dexterous robotic hands. Our key idea is to treat affordance regions as semantic constraints, guiding the agent toward meaningful grasps and manipulation behaviors.

\begin{figure}[!htb]
    \centering
    \includegraphics[width=\linewidth]{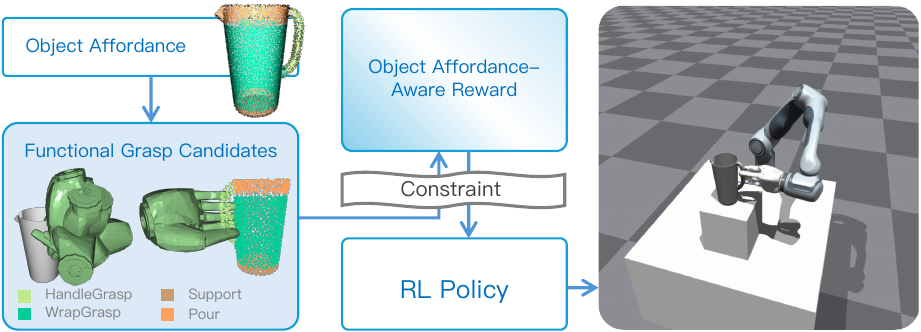}
    \caption{Overview of the proposed object affordance-guided reinforcement learning framework. Object affordance maps are used to generate semantically meaningful functional grasp candidates. These candidates serve as constraints and priors to guide the RL policy, which is further optimized using an object affordance-aware reward. The resulting policy enables dexterous manipulation that is functionally grounded and task-relevant.}
    \label{fig.heading_photo}
\end{figure}

To this end, we construct a grasp synthesis framework that generates affordance-aligned grasp candidates based on contact semantic maps and a novel voting-based classification algorithm. These grasp poses are then embedded into a generalizable RL pipeline, which includes motion-feasibility filtering, sub-task-aware policy transitions, and an affordance-aware reward formulation. By constraining the learning process to functionally relevant regions of the object, our method accelerates training convergence and improves task success.

We evaluate our approach on a suite of dexterous manipulation tasks using a simulated multi-fingered robotic hand. Tasks include grasping and lifting a cube, functionally grasping a jug by its handle, and reorienting a hammer for tool use. Experimental results show that our framework consistently outperforms baseline methods, yielding up to 27.7\% improvement in grasp success rate for semantically guided tasks and an average of 15.4\% improvement in overall task success.

In summary, our contributions are threefold:
\begin{itemize}
\item \textbf{An object affordance-guided RL framework:}
We present a reinforcement learning (RL) pipeline that incorporates functional grasp candidates derived from object affordance maps as semantic constraints. These candidates are integrated into the policy learning process through motion-feasibility filtering and a structured sub-task transition mechanism, enabling goal-aware and physically feasible manipulation behavior.
\item \textbf{Object affordance-aware reward design:} We propose a generalizable reward function that explicitly encodes both object affordance guidance and task-specific objectives. The reward formulation combines an affordance-centric proximity term with sub-task-aligned goals, such as lifting or in-hand re-orientation, effectively shaping the agent’s behavior toward functionally meaningful interactions.

\item \textbf{Empirical validation with improved efficiency and performance:} We demonstrate the effectiveness of our approach across multiple dexterous manipulation tasks involving varied object geometries and functional requirements. Experimental results show that our method consistently outperforms affordance-agnostic baselines, achieving higher success rates and substantially faster convergence during training.

\end{itemize}

\section{Related Work}
\label{sec:relatedwork}
\subsection{Dexterous Multi-Fingered Robotic Manipulation RL}

Reinforcement learning (RL) has emerged as a promising approach for learning control policies in high-dimensional dexterous manipulation tasks~\cite{andrychowicz2020learning, chen2022humanlevelbimanualdexterousmanipulation,chen2023sequential,chen2023bi}. Multi-fingered robotic hands offer flexibility akin to human hands, yet present significant challenges due to the complexity of joint coordination, contact dynamics, and delayed reward structures. Several works have explored model-free RL for dexterous in-hand manipulation~\cite{andrychowicz2020learning}, leveraging dense proprioceptive signals and domain randomization to facilitate sim-to-real transfer. However, many of these policies lack semantic understanding and often fail to generalize across diverse objects or tasks. Unlike humans, RL agents often resort to trial-and-error exploration, ignoring functional priors inherent in object geometry and use. Our work addresses this limitation by incorporating object affordance as a semantic constraint, thus bridging the gap between low-level motor learning and high-level object-centric reasoning.

\subsection{Object Affordance in Robotic Learning}

The concept of object affordance, originally proposed in psychology~\cite{gibson1979ecological}, has been widely adopted in robotics to represent the potential actions associated with object regions. Prior work has focused on affordance detection using visual cues~\cite{do2018affordancenet, nguyen2017object,zeng2022robotic,wu2023learning,mandikal2021learning,wang2024tooleenet}, where deep neural networks are trained to segment or classify object parts according to functional categories such as "grasp", "push", or "open". Recently, researchers have begun to integrate affordance signals into policy learning itself, giving rise to object affordance-guided reinforcement learning~\cite{lee2024affordance, borja2022affordance}. These approaches utilize affordance maps as priors to guide exploration, constrain action spaces, or shape reward functions, thereby improving learning efficiency and generalization. For instance, Lee et al.\cite{lee2024affordance} propose a framework that uses visual affordances to narrow down manipulation strategies, while Borja et al.\cite{borja2022affordance} demonstrate that leveraging affordance knowledge enables more sample-efficient policy learning in task-conditioned settings. However, these methods are primarily designed for simplified manipulation tasks or parallel-jaw grippers and often do not generalize well to complex, high-DoF dexterous hands.

In contrast, our approach directly incorporates object affordance into both grasp pose generation and dexterous policy optimization, enabling affordance-driven constraint filtering and reward shaping within a unified multi-stage reinforcement learning framework.

\subsection{Functional Grasp Pose Generation}
Functional grasping focuses on achieving grasps that not only ensure physical stability but also enable downstream task execution~\cite{montesano2008learning, cardellicchio2013grasping,wu2024cross}. Unlike traditional force-closure grasps, functional grasps consider semantic intent—e.g., grasping a mug by the handle to pour. Several studies have investigated learning-based grasp synthesis with semantic grounding~\cite{patankar2023task,mandikal2021learning}. 
Optimization-based grasp generation techniques have also incorporated hand-object interaction information to propose grasp poses with contact information~\cite{zhang2024multi}. Our work extends this line by proposing a voting-based functional grasp classifier that fuses contact semantics and affordance labels, and by integrating constraints into RL-based manipulation training.

\subsection{Constraint-Aware Reinforcement Learning}
Incorporating constraints into reinforcement learning is critical for safe and efficient policy learning~\cite{achiam2017constrained,turchetta2016safe,berkenkamp2017safe}, especially in robotics. 
In the context of manipulation, constraints are often introduced through motion planning or physical feasibility~\cite{song2020grasping}. However, semantic constraints derived from object structure—such as affordance-aligned grasping—have been largely underexplored. Our method integrates affordance-informed grasp candidates as hard constraints in policy optimization and augments the reward function with semantic priors, promoting efficient and goal-aligned exploration in complex manipulation tasks.

\begin{figure*}[!htb]
    \centering
    \includegraphics[width=\linewidth]{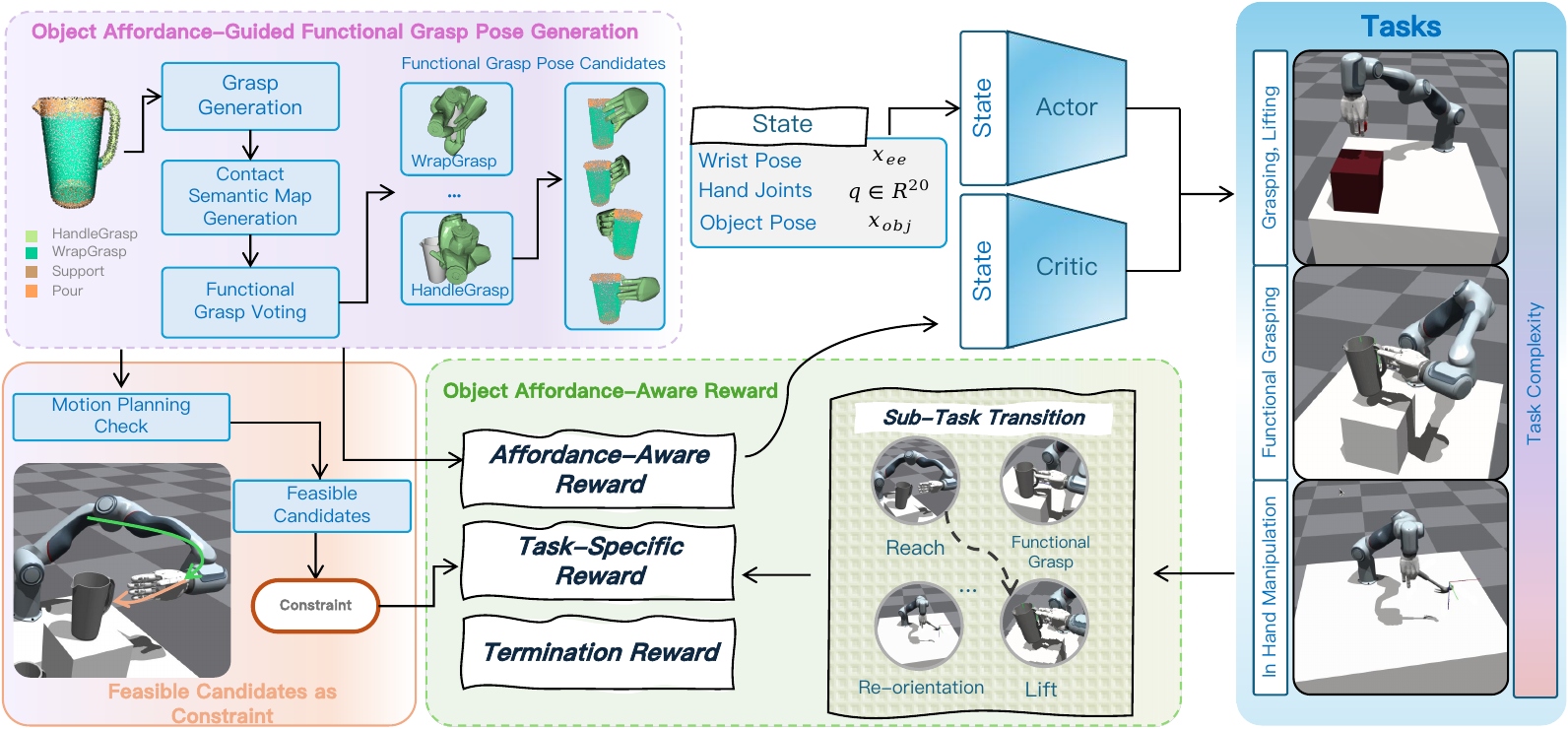}
    \caption{Overview of the object affordance-guided reinforcement learning pipeline for various dexterous manipulation tasks.
Functional grasp candidates are first generated based on object affordance information. Feasible candidates are then filtered through motion planning checks and used as constraints during RL training. Additionally, an affordance-aware reward is designed to guide policy learning across tasks such as pick-and-place, functional grasping, and in-hand reorientation.}
    \label{fig.pipeline}
\end{figure*}

\section{Problem Statement and Methods}\label{sec:method}
\subsection{Problem Formulation}

The design of dexterous hands aims to replicate the highly sophisticated manipulation capabilities of human hands. To achieve this, an intelligent agent must not only precisely control the high degrees of freedom (DoFs) of the hand joints, but also exhibit adaptive strategies informed by object affordance information. For instance, when manipulating a pitcher, a dexterous hand typically interacts with specific affordance regions of the object that facilitate effective handling.

We propose a reinforcement learning (RL) framework for dexterous manipulation that is guided by object affordances, as shown in Fig.~\ref{fig.pipeline}. Our goal is to enhance the learning efficiency of human-like manipulation behaviors. The proposed method consists of three key components: object affordance-guided functional grasp constraint generation, and object affordance-aware dexterous RL training pipeline and corresponding reward design. To this end, we formally define the problem as a Markov Decision Process (MDP) augmented with object affordance information.

\textbf{State Space} The state $\mathbf{s} \in \mathcal{S}$ includes:
\begin{itemize}
    \item     The proprioceptive state of the hand, including joint angles, joint velocities and finger tip poses,
    \item The proprioceptive state of robotic arm, including wrist pose and velocity,
    \item Pose of manipulated object, object affordance map,
    \item Functional grasp pose candidates generated based on object affordance,
    \item sub-task stage.
\end{itemize}

\textbf{Action Space} The action $\mathbf{a} \in \mathcal{A}$ corresponds to control commands for the hand's joints and robotic arm's wrist pose.

\textbf{Transition Dynamics} $p(s_{t+1} \mid s_t, a_t)$ are determined by the underlying physics simulation environment.

\textbf{Object Affordance Guidance} We assume each object is associated with an object affordance map $\mathcal{A}_o$, highlighting key regions such as handles. Functional grasp poses are generated based upon object affordance map, as detailed in Sec.~\ref{subsection:grasp_generation}. The object affordance map and the generated functional grasp poses are integrated into the reinforcement learning process to guide the policy's exploration and improve training speed, as introduced in Sec.~\ref{subsection:object_affordance_guided_RL_training}.

\textbf{Reward Function} The object affordance-aware reward function $r(\mathbf{s}, \mathbf{a})$ is designed to encourage the agent to interact with functionally-relevant regions of the object, and complete the RL training efficiently, as described in Sec~\ref{subsection:reward_design}.

\textbf{Learning Objective} The agent learns a policy $\pi(\mathbf{a} \mid \mathbf{s})$ that maximizes the expected cumulative discounted reward while satisfying task-relevant constraints, where $\gamma \in [0, 1)$ is the discount factor. Specifically, we incorporate constraints derived from the object's affordance and the predicted functional grasp pose, which guide the policy toward feasible and semantically meaningful grasps.

\subsection{Object Affordance-Guided Functional Grasping Pose Generation}
\label{subsection:grasp_generation}

To generate functional grasping poses, we propose an object affordance-guided grasp synthesis framework. First, building upon our previous work~\cite{zhang2024multi}, we generate dexterous robotic hand grasping data conditioned on object models. This dataset includes grasping poses and contact semantic maps. Subsequently, we introduce a voting-based functional grasp classification method, which leverages both the contact semantic maps and the object affordance maps to predict grasp categories.

\textbf{Grasp Pose Optimization} In grasp generation, robotic hand wrist poses and joint poses are optimized by minimizing the energy function 
$E = E_{\text{DFC}} + E_{\text{HO}} + E_{\text{Robot}}$, where $E_{\text{DFC}}$ represents the grasp quality approximated by a differential force closure (DFC) estimator. $E_{\text{HO}}$ and $E_{\text{Robot}}$ correspond to the hand-object interaction energy and robotic constraint-associated energy~\cite{zhang2024multi}.

DFC estimator is employed for assessing grasp quality by estimating wrench disregarding frictional forces. The score $E_{\rm DFC}$ is derived using the contact normal vectors $c$.
\begin{equation}
\begin{gathered}
E_{\rm DFC}=\|G c\|^2 \\
G=\left[\begin{array}{ccc}
I_3 & \cdots & I_3 \\
{\left[\psi_1\right]_{\times}} & \cdots & {\left[\psi_n\right]_{\times}}
\end{array}\right] \\
{\left[\psi_k\right]_{\times}=\left[\begin{array}{ccc}
0 & -\psi_k^{(z)} & \psi_k^{(y)} \\
\psi_k^{(z)} & 0 & -\psi_k^{(x)} \\
-\psi_k^{(y)} & \psi_k^{(x)} & 0
\end{array}\right]}
\end{gathered}
\end{equation}
where, $\Psi=\{\psi_1, \cdots,\psi_n\}$ represents the set of contact point candidates, sampled from hand surface, term $c \in \mathbb{R}^{n \times 3}$ denotes the normals of object surface at the contact points in~$\Psi$, and $n$ indicates the number of contact points.

\textbf{Contact Semantic Map} The contact semantic map, denoted as $M_{\text{CoS}} \in \mathbb{R}^{P}$, is constructed by computing the distance between each joint or link of the robotic hand and the surface points of the object. $P$ is the number of points on the object surface. For each point $p_i$, the corresponding label is defined as: 
\begin{equation}
M_{\text{CoS}}(p_i) = l_j, \quad l_j \in \mathcal{L}_{\text{hand}}
\end{equation}
where $L_{\rm hand}$=\{no contact, palm, thumb, index, middle, ring,pinky\} represents the set of hand parts. This label specifies which part of the robotic hand—such as a specific finger or the palm—is in contact with point $p_i$ under a given grasping pose.

\textbf{Object Affordance Map} The object affordance map $M_{\text{Aff}} \in \mathbb{R}^{P}$ assigns an affordance label to each surface point of the object:
\begin{equation} 
M_{\text{Aff}}(p_i) = a_k, \quad a_k \in { \text{predefined affordance classes}} 
\end{equation}
Each affordance label $a_k$ represents a functional category associated with the corresponding object point, providing critical cues for functional grasp generation.

\textbf{Voting-Based Grasp Classification } To classify the functional category of a grasp, we propose a voting-based algorithm that integrates contact semantics and object affordance information. For each contact point $p_i$ between the robotic hand and the object surface, we first retrieve its $N$ nearest affordance labels from the object affordance map:

\begin{equation}
\mathcal{N}(p_i) = \{ a_k^{(1)}, a_k^{(2)}, \dots, a_k^{(N)} \},
\end{equation}

where each $a_k^{(j)}$ denotes the affordance class of a nearby point on the object surface. These retrieved labels reflect the local functional context around each contact point.

Next, for each finger $f_m$ of the robotic hand, we collect all its associated contact points $\mathcal{P}_{f_m}$ and perform a majority vote over their nearest neighbor affordance labels. This results in a finger-level predicted affordance class:

\begin{equation}
A_{f_m} = \operatorname{mode} \left( \bigcup_{p_i \in \mathcal{P}_{f_m}} \mathcal{N}(p_i) \right),
\end{equation}

where $\operatorname{mode}(\cdot)$ denotes the most frequent affordance label among the finger’s contact points.

Finally, the overall grasp-level affordance classification $A_{\text{grasp}}$ is computed by aggregating the predictions from all fingers using a weighted voting scheme. Fingers that play a more critical role in grasp stability or functionality are assigned higher weights $w_{f_m}$. The grasp-level classification is then given by:

\begin{equation}
A_{\text{grasp}} = \operatorname{argmax}_a \left( \sum_{m=1}^{M} w_{f_m} \cdot \mathbb{I}(A_{f_m} = a) \right),
\end{equation}

where $M$ is the total number of fingers, and $\mathbb{I}(\cdot)$ is the indicator function that equals 1 when the predicted label matches $a$, and 0 otherwise. This weighted voting strategy allows the system to emphasize more reliable contact regions in determining the final affordance classification of the grasp.

\subsection{Generalizable Object Affordance Guided-RL Pipeline}
\label{subsection:object_affordance_guided_RL_training}
We design a generalizable reinforcement learning (RL) pipeline that leverages functional grasp poses guided by object affordances to constrain and enhance the training process of manipulation policies. Following the object affordance-guided grasp synthesis stage, we obtain a set of grasp pose candidates $
\mathcal{G} = \{g_1, g_2, \dots, g_n\}
$, each associated with a grasp type $t_j \in \mathcal{T}$. These affordance-labeled grasp candidates serve as structured priors and constraints to guide the subsequent RL training. The RL pipeline is composed of three key components: (1) Sub-task Transition, (2) Motion-Feasibility Filtering, and (3) Reward Design, with the third module elaborated in the next subsection.

\textbf{Sub-task Transition} First, the Sub-task Transition module monitors the environment state $s_t \in \mathcal{S}_t
$ to determine whether the current sub-task $\tau_i \in \mathcal{T}_{\text{sub}}
$ has been successfully completed. Upon reaching a terminal condition $\phi(s_t) = \text{True}
$, the pipeline transitions to the next sub-task. This mechanism also controls the dynamic switching of reward functions, enabling structured learning in complex multi-stage manipulation tasks.

\textbf{Motion-Feasibility Filtering} Second, the Motion-Feasibility Filtering module ensures that only kinematically reachable and semantically meaningful grasp candidates are retained for policy optimization. Given a set of grasp candidates $mathcal{G}$, we evaluate each $g_i$ by performing a motion planning feasibility check:

\begin{equation}
f_{\text{reach}}(g_i, s_t) = 
\begin{cases}
1, & \text{if reachable without collision} \\
0, & \text{otherwise}
\end{cases}
\end{equation}

Only candidates satisfying $f_{\text{reach}}(g_i, s_t) = 1
$ are included in the training set. If none of the candidates are reachable under the current scene configuration, the environment is reset and the reward weight is set to zero for this episode to penalize infeasible initialization.

The filtered candidates $\mathcal{G}_{\text{valid}} \subset \mathcal{G}
$ are then used as constraints to regularize the RL policy search, encouraging the agent to operate within a semantically and physically valid action space. These constraints are later integrated into the reward formulation, as detailed in Section~\ref{subsection:reward_design}.

\textbf{Policy Learning} For policy optimization, we adopt the Soft Actor-Critic (SAC) algorithm, a state-of-the-art off-policy reinforcement learning method known for its sample efficiency and stability in continuous action spaces. The actor-critic architecture is trained using observations that include the robot wrist pose \( \mathbf{x}_{\text{ee}} \), hand joint configuration \( \mathbf{q} \in \mathbb{R}^{20} \), and the object pose \( \mathbf{x}_{\text{obj}} \), as illustrated in Fig.~\ref{fig.pipeline}.

During training, the actor network estimates the wrist delta pose and hand joint pose. The robotic arm is controlled in Cartesian space based on estimated delta pose \( \Delta \mathbf{x}_{\text{ee}} \), which is then converted to joint velocities using Jacobian-based inverse kinematics:

\begin{equation}
\Delta \mathbf{q}_{\text{arm}} = J^\dagger(\mathbf{q}) \cdot \Delta \mathbf{x}_{\text{ee}}
\end{equation}

\noindent
where \( J^\dagger(\mathbf{q}) \) is the pseudo-inverse of the Jacobian matrix evaluated at the current joint configuration.

In contrast, the robotic hand is controlled directly in joint space. The policy outputs a target joint position vector \( \mathbf{q}_{\text{hand}}^{\text{target}} \in \mathbb{R}^{20} \).

\subsection{Object Affordance-Aware Reward Design}
\label{subsection:reward_design}

We design an object affordance-aware reward to ensures that the RL agent learns task-relevant behaviors that are semantically meaningful and physically executable. To unify reward design across diverse manipulation tasks, we introduce a generalized reward function schema:

\begin{equation}
R(s, a) = w_a \cdot R_{\text{affordance}} + w_t \cdot R_{\text{task}} - w_p \cdot R_{\text{penalty}}
\end{equation}

\noindent
Where:
\begin{itemize}
  \item $R_{\text{affordance}}$: Reward for object affordance-aware manipulation.
  \item $R_{\text{task}}$: Task-specific reward (e.g., position, orientation, force).
  \item $R_{\text{penalty}}$: Penalty for excessive or unstable actions.
  \item $w_a$, $w_t$, $w_p$: Task-specific weighting coefficients.
\end{itemize}

This formulation allows for task-adaptive reward design while preserving a consistent structure. We instantiate this schema across multiple functional manipulation tasks (e.g., \textit{Jug Grasping and Lifting}, \textit{Hammer Grasping and Re-Orientation}), each with customized reward components derived from the same principle of object affordance-aware and functional grasp candidate-constrained learning.

\subsubsection{\textbf{Object Affordance-Aware Manipulation Reward $R_{\text{affordance}}$}}

To encourage stable and functional manipulation poses aligned with the object's affordance regions, we define the object affordance-aware manipulation reward as an exponential function of the distance between each fingertip and the target affordance region:

\begin{equation}
R_{\text{affordance}} = \exp\left(-\alpha \sum_{i=1}^{N_f} \| p_{\text{tip}_i} - p_{\text{afford}} \|_2 \right)
\end{equation}

Here, $N_f$ denotes the number of fingertips, and $p_\text{afford}$ represents the position of the semantically meaningful region identified by the object affordance map. $p_\text{tip}$ denotes the position of finger tip. The weight $\alpha$ controls the spatial sensitivity.

This term ensures that the agent prioritizes contact with functional regions of the object, effectively embedding the affordance prior into the manipulation behavior.

\subsubsection{\textbf{Task-Specific Reward $R_{\text{task}}$}}

The task reward captures the functional objective of each manipulation task, such as grasping, lifting, or in-hand re-orienting. It is constructed using task-dependent metrics:

\paragraph{Sub Task 1: Grasping}
\begin{equation}
R_{\text{task}}^{\text{grasp}} = \exp\left(z_{\text{current}} - z_{\text{feasible candidate}} \right)
\end{equation}

\paragraph{Sub Task 2: Lifting}
\begin{equation}
R_{\text{task}}^{\text{lift}} = \exp\left(z_{\text{current}} - z_{\text{initial}} \right)
\end{equation}

\paragraph{Sub Task 3: In Hand Re-Orientation}
\begin{equation}
R_{\text{task}}^{\text{orient}} = -2 \cdot \arcsin\left(2 \cdot \langle r_{\text{current}}, r_{\text{target}} \rangle^2 - 1\right)
\end{equation}

The generated grasp poses primarily influence the reward computation for grasp-related sub-tasks. For lifting tasks, the target pose is derived based on a reliable grasp configuration that ensures stability and kinematic feasibility. In in-hand re-orientation tasks, the orientation target is defined with respect to the affordance frame, ensuring that the desired rotation aligns with the intended functional use of the object.

By embedding these constraint-informed objectives into the task-specific reward, we effectively guide each skill to operate within semantically valid and physically meaningful boundaries. Grounding the reward signals in task semantics ensures that the agent acquires behaviors aligned with the object’s intended use, as implicitly defined by its affordance structure.

    \begin{figure}[!ht]
    \centering
    \begin{subfigure}[t]{0.15\textwidth}
        \centering
        \includegraphics[width=0.8\linewidth]{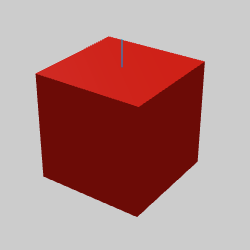}
        \caption{Cube}
        \label{fig:cube}
    \end{subfigure}
        \begin{subfigure}[t]{0.15\textwidth}
        \centering
        \includegraphics[width=0.8\linewidth]{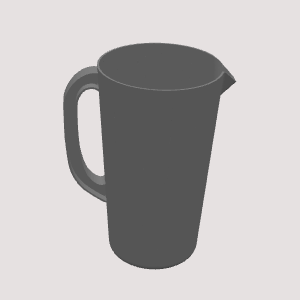}
        \caption{Jug}
        \label{fig:jug}
    \end{subfigure}
    \begin{subfigure}[t]{0.15\textwidth}
        \centering
        \includegraphics[width=0.8\linewidth]{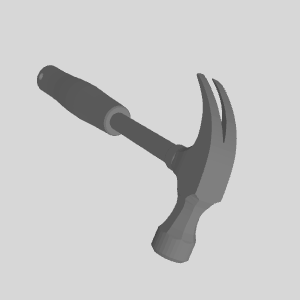}
        \caption{Hammer}
        \label{fig:hammer}
    \end{subfigure}
    \caption{Examples of object affordances used in the task.}
    \label{fig:affordance_objects}
\end{figure}

\begin{table}
\caption{Task descriptions and scenes in experiments. Tasks are listed in order of difficulty concerning control and finger dexterity.}
\begin{tabular}{ | m{5em} | m{2.6cm}| m{3cm} | }
  \hline
  Task & Task Description& Illustration \\ 
  \hline
  Task1:~Grasp and Lift & Grasp a cube and lift it above a certain distance & \raisebox{-0.1\totalheight}{\includegraphics[width=0.16\textwidth]{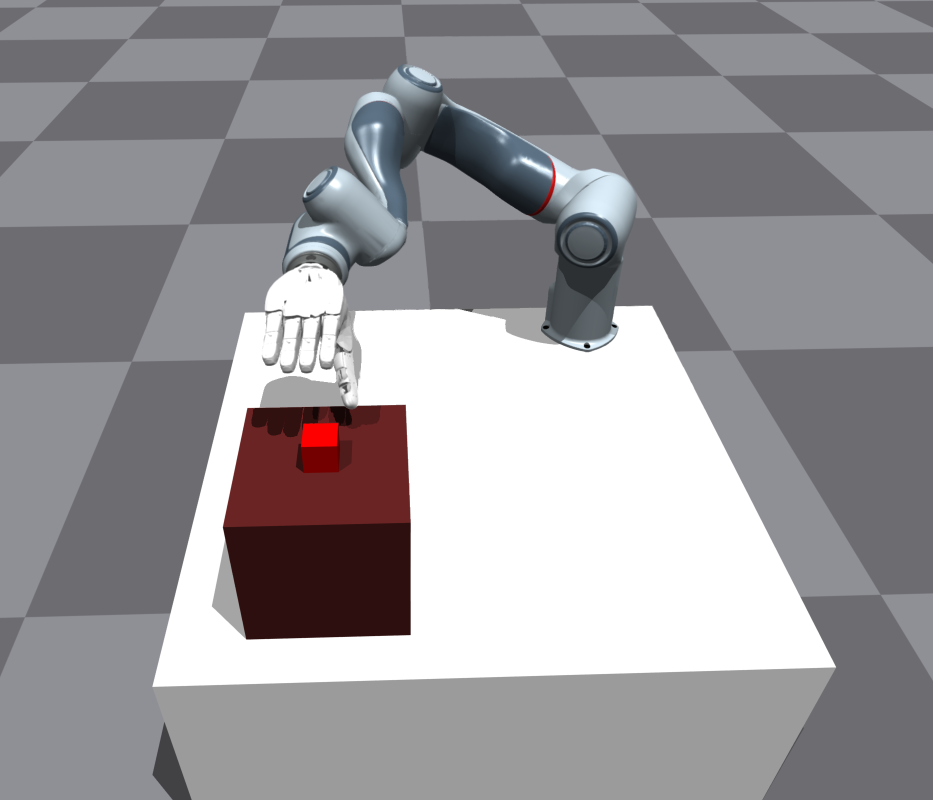}} \\ 
  \hline
    Task 2: Jug Functional Grasp and Lift & Grasp a jug with grasp type of HandleGrasp, and lifting& \raisebox{-0.1\totalheight}{\includegraphics[width=0.16\textwidth]{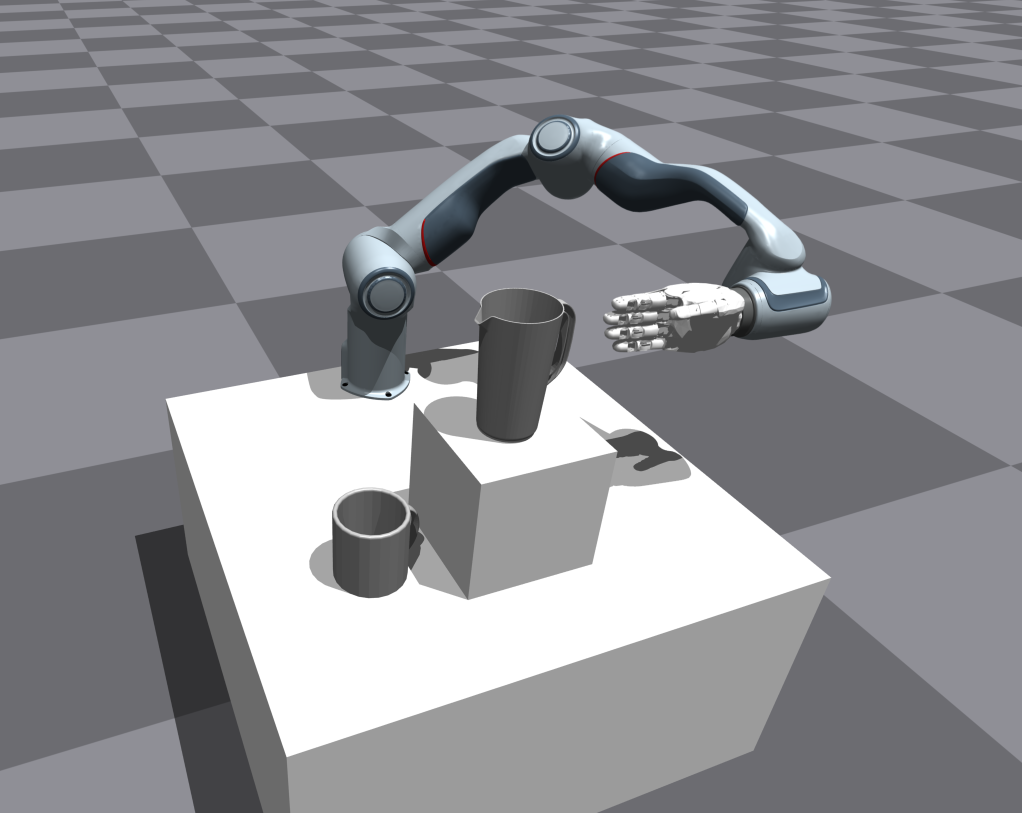}} \\
  \hline
Task 3: Hammer Use & hold a falling hammer, manipulate it into a stable position, reorient and use it on a stationary nail & \raisebox{-0.1\totalheight}{\includegraphics[width=0.16\textwidth]{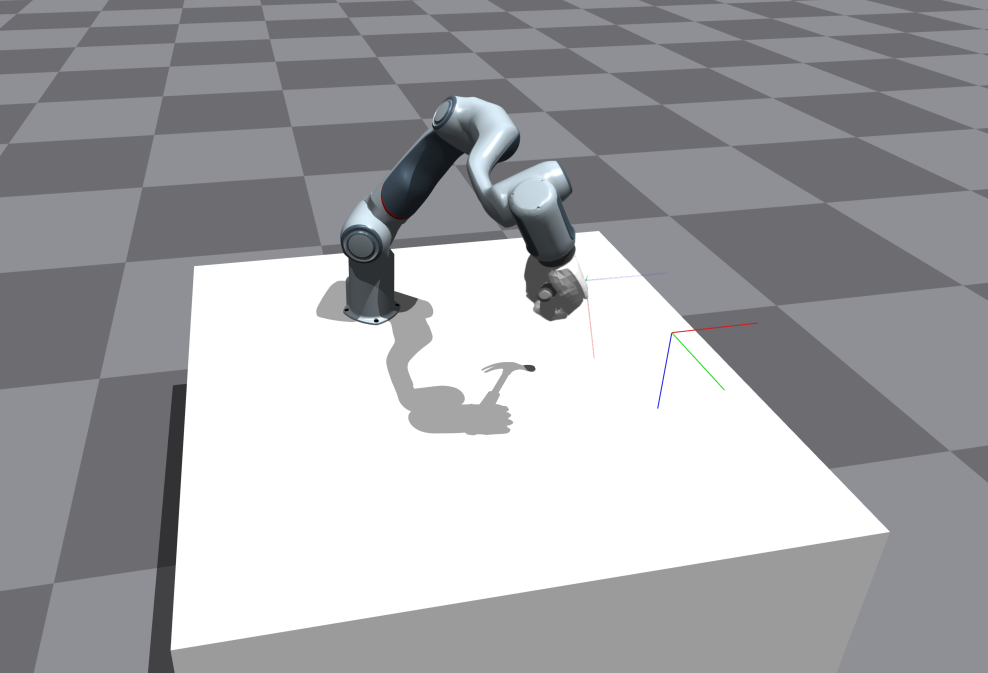}} \\ 
  \hline
\end{tabular}
\label{tab:task_table}

\end{table}

\subsubsection{Action Penalty $R_{\text{penalty}}$}

To discourage inefficient or unstable control, we apply a quadratic penalty on the magnitude of actions:

\begin{equation}
R_{\text{penalty}} = \sum_{i} \left( \text{action}_i^2 \right)
\end{equation}

This term stabilizes training and prevents unnecessary movement or energy consumption. While not directly linked to object affordance, it acts as a regularizer to support smooth policy learning.

\setlength{\textfloatsep}{5pt}

\section{Experiment}
\label{sec:experiment}
\subsection{Experimental Setup}

Our robotic simulation environment utilizes the Diana7 robotic arm equipped with the DLR-HIT Multi-Fingered Robotic Hand~\cite{liu2008multisensory} based on Isaac Gym~\cite{makoviychuk2021isaac}. The experimental setup includes three representative objects: a cube, a jug, and a hammer, as shown in Fig.~\ref{fig:affordance_objects}. 
Our task suite includes basic grasping and lifting of a cube, jug manipulation including functional grasping and lifting, and tool-use with a hammer from grasping to re-orientation. Detailed task descriptions and corresponding simulation scenes are illustrated in Tab.~\ref{tab:task_table}.

\begin{figure*}[ht]
\centering
\begin{subfigure}{.23\textwidth}
  \centering
  \includegraphics[width=\linewidth]{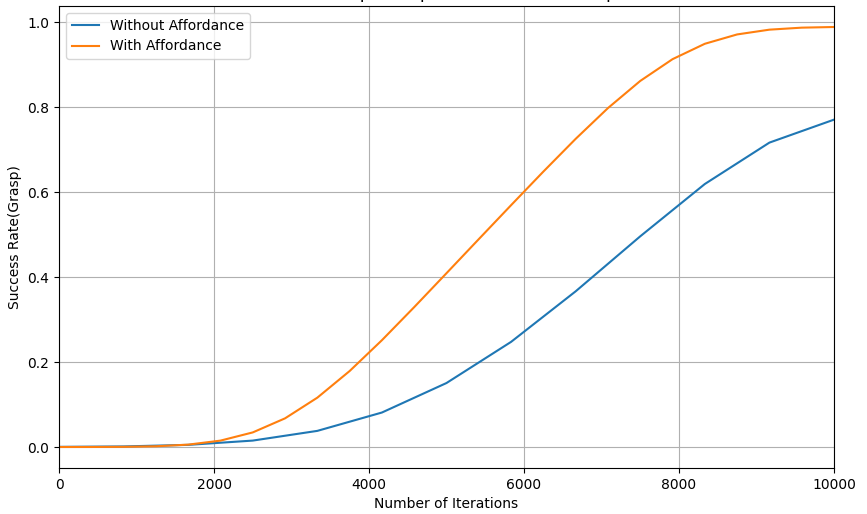}
  \caption{Task 1: Grasp SR}
  \label{fig:cube_grasp_ppo_without}
\end{subfigure}
\begin{subfigure}{.23\textwidth}
  \centering
  \includegraphics[width=\linewidth]{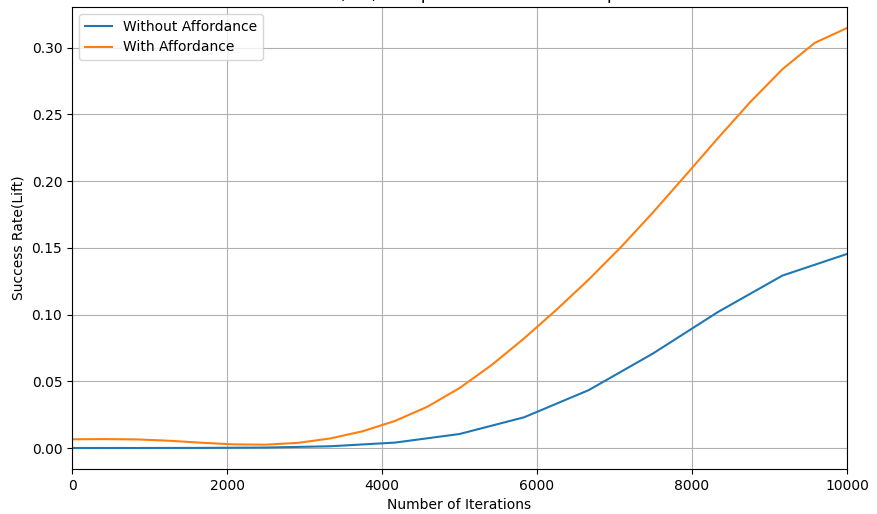}
  \caption{Task 1: Lifting SR}
  \label{fig:cube_lift_ppo_without}
\end{subfigure}
\begin{subfigure}{.23\textwidth}
  \centering
  \includegraphics[width=\linewidth]{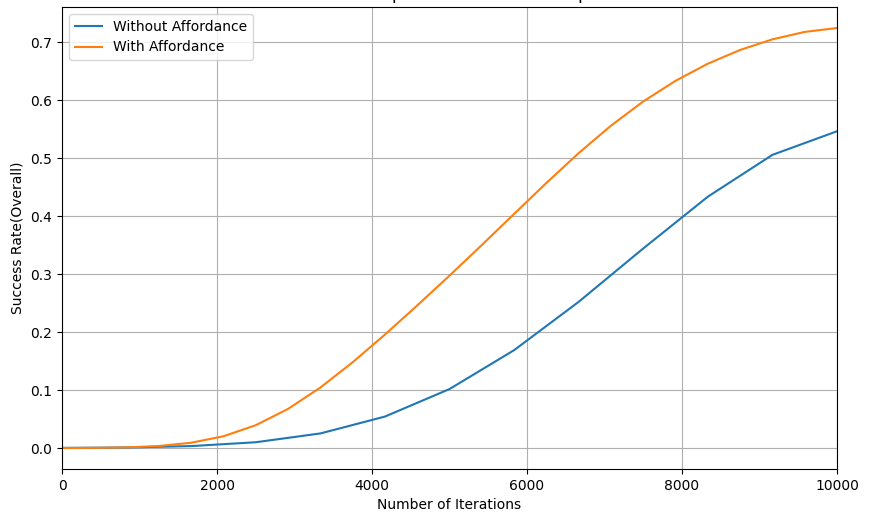}
  \caption{Task 1: Overall SR}
  \label{fig:cube_episode_ppo_without}
\end{subfigure}
\begin{subfigure}{.23\textwidth}
  \centering
  \includegraphics[width=\linewidth]{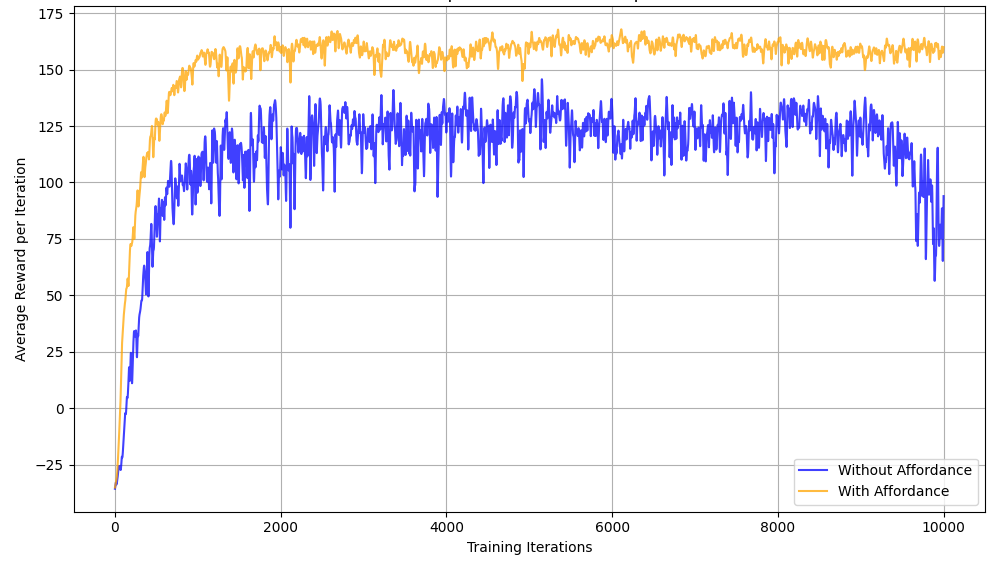}
  \caption{Task 1: Reward}
  \label{fig:cube_reward_ppo_without}
\end{subfigure}

\vspace{1em}

\begin{subfigure}{.23\textwidth}
  \centering
  \includegraphics[width=\linewidth]{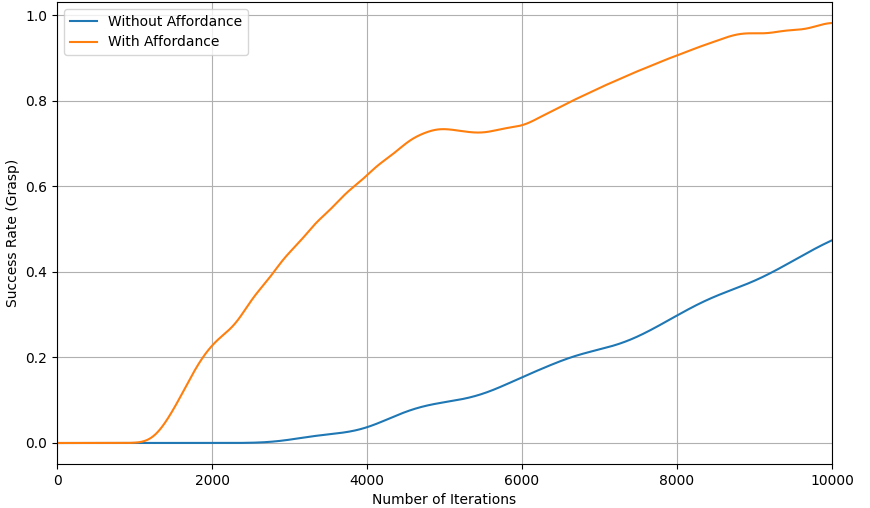}
  \caption{Task 2: Grasp SR}
  \label{fig:jug_grasp_sac}
\end{subfigure}
\begin{subfigure}{.23\textwidth}
  \centering
  \includegraphics[width=\linewidth]{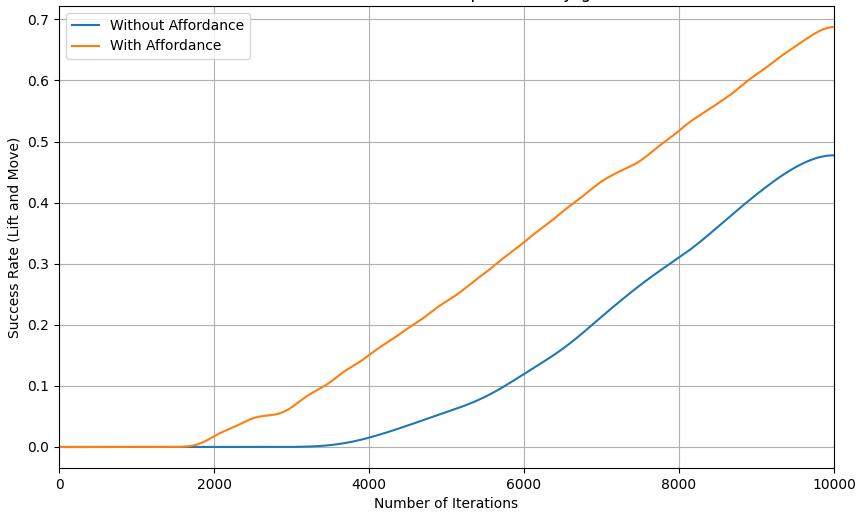}
  \caption{Task 2: Lifting SR}
  \label{fig:jug_lift_sac}
\end{subfigure}
\begin{subfigure}{.23\textwidth}
  \centering
  \includegraphics[width=\linewidth]{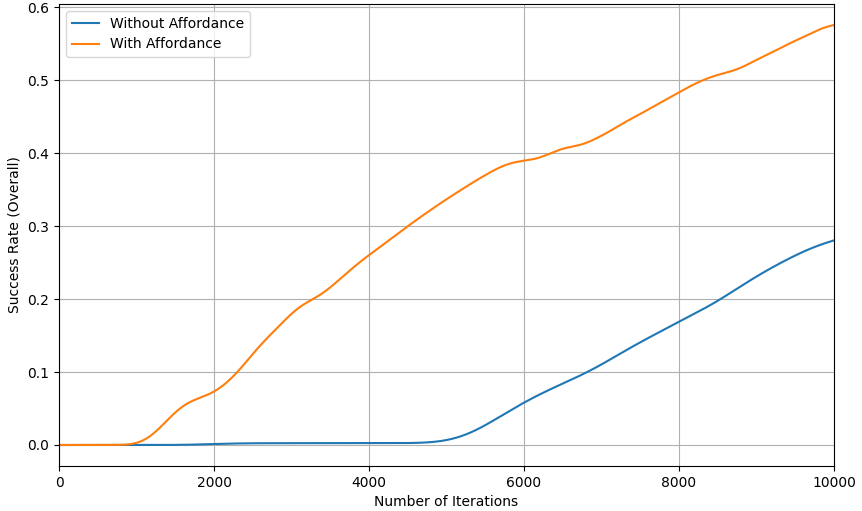}
  \caption{Task 2: Overall SR}
  \label{fig:jug_lift_overall}
\end{subfigure}
\begin{subfigure}{.23\textwidth}
  \centering
  \includegraphics[width=\linewidth]{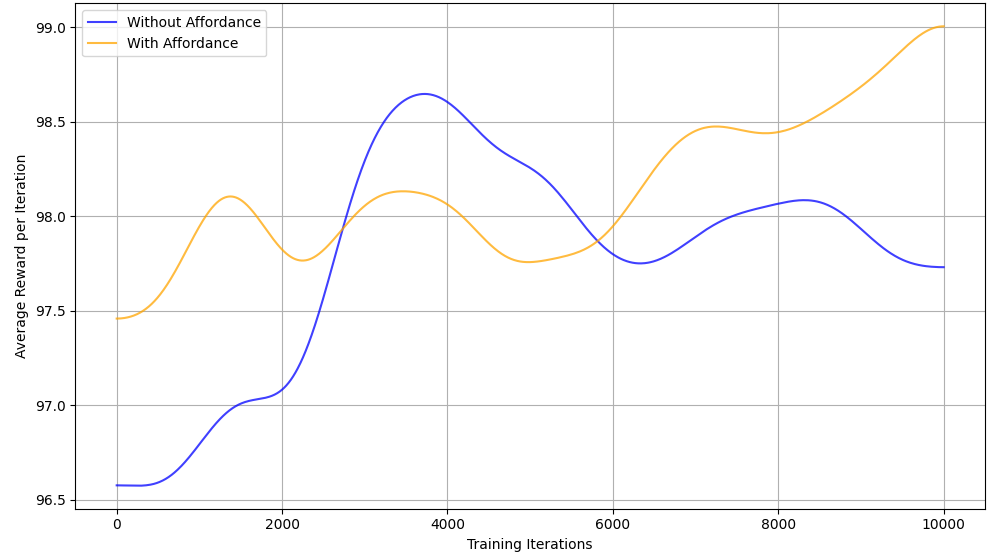}
  \caption{Task 2: Reward}
  \label{fig:jug_reward_curve}
\end{subfigure}

\vspace{1em}

\begin{subfigure}{.23\textwidth}
  \centering
  \includegraphics[width=\linewidth]{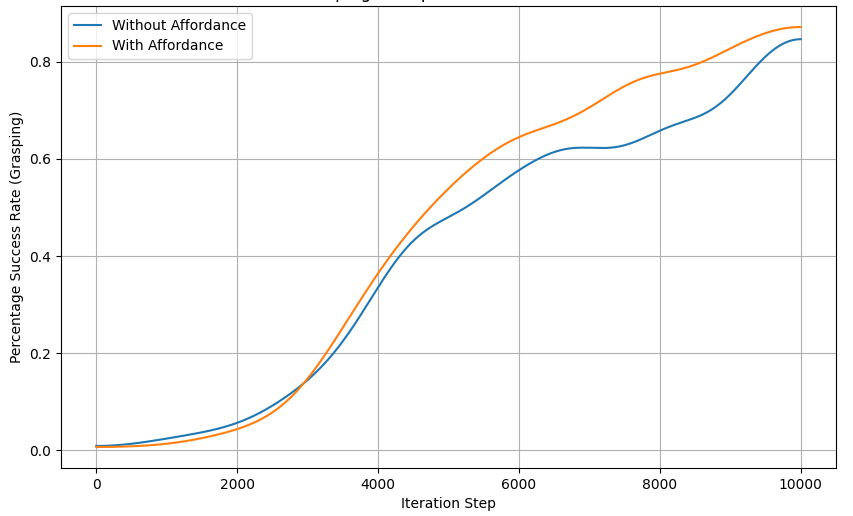}
  \caption{Task 3: Grasp SR}
  \label{fig:hammer_grasp_ppo_sr}
\end{subfigure}
\begin{subfigure}{.23\textwidth}
  \centering
  \includegraphics[width=\linewidth]{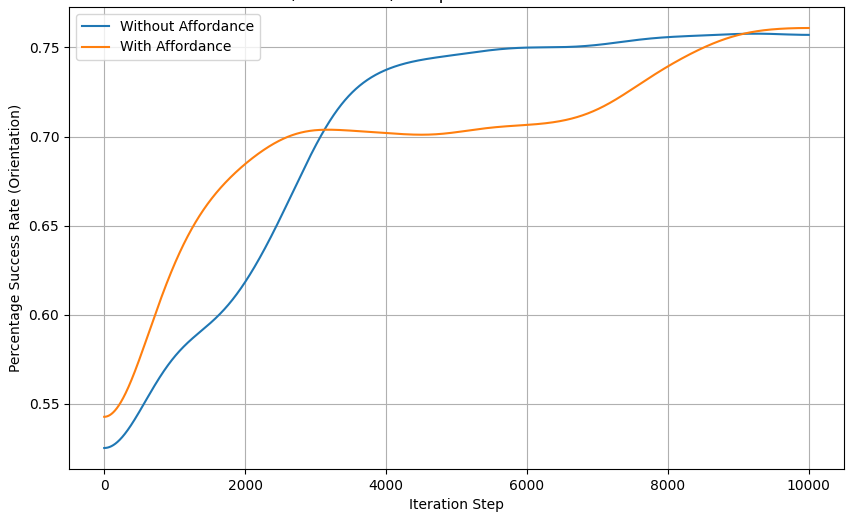}
  \caption{Task 3: Orient SR}
  \label{fig:hammer_orient_ppo_sr}
\end{subfigure}
\begin{subfigure}{.23\textwidth}
  \centering
  \includegraphics[width=\linewidth]{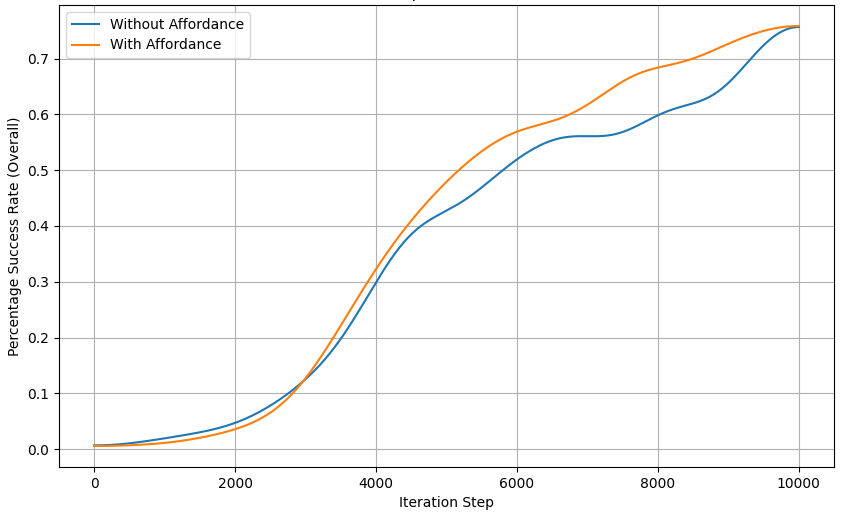}
  \caption{Task 3: Overall SR}
  \label{fig:hammer_overall_ppo_sr}
\end{subfigure}
\begin{subfigure}{.23\textwidth}
  \centering
  \includegraphics[width=\linewidth]{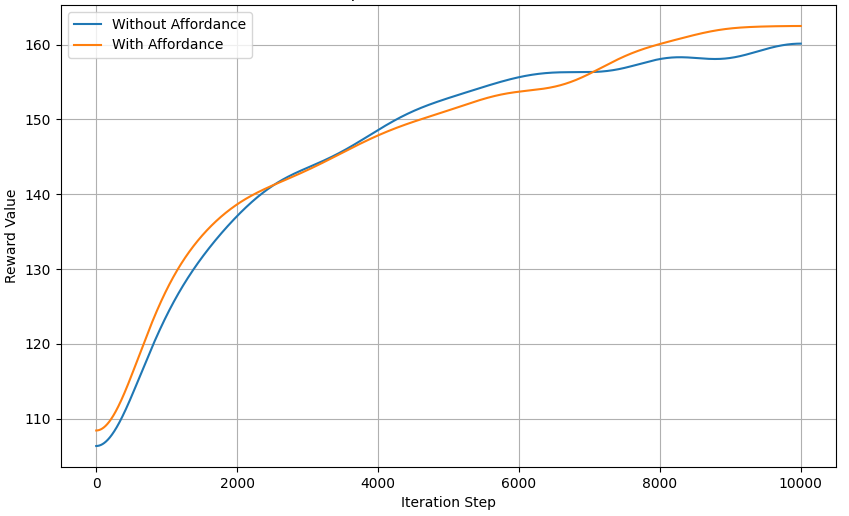}
  \caption{Task 3: Reward}
  \label{fig:hammer_reward_ppo}
\end{subfigure}


\caption{Performance comparison across three tasks based on optimal policy. Task 1: Cube grasping and lifting, Task 2: Jug functional grasping and lifting, Task 3: Hammer Grasping and Re-Orientation.}
\label{fig:All_Policy_Comparison}
\end{figure*}

\subsection{Comparative Experiments}

To compare the performance of our proposed approach under different policies and training paradigms, we design a set of ablation and baseline experiments including:
\begin{itemize}
    \item PPO~\cite{schulman2017proximal} without object affordance guidance.
    \item PPO~\cite{schulman2017proximal} with object affordance guidance.
    \item SAC~\cite{haarnoja2018soft} without object affordance guidance.
    \item SAC~\cite{haarnoja2018soft} with object affordance guidance.
\end{itemize}
All experiments are conducted using an NVIDIA RTX 4090 GPU. Training is performed for up to 10,000 epochs to ensure sufficient convergence across tasks. We use a batch size of 32, which strikes a practical balance between gradient estimate stability and memory efficiency. This batch size is commonly adopted in continuous control settings, where it provides reliable learning signals while mitigating the high variance often associated with smaller batches~\cite{chen2022humanlevelbimanualdexterousmanipulation}.

\subsection{Evaluation Metrics}
We evaluate the performance of different algorithms across manipulation tasks using the following metrics:
\begin{itemize}
    \item     Overall Success Rate (Overall SR): The percentage of environments in which the task was successfully completed. Results are averaged over three random seeds and smoothed using a Gaussian filter with a sigma of 200.

\end{itemize}

In addition to overall task success, we also report performance on sub-tasks to gain finer-grained insights:
\begin{itemize}
    \item     Grasping Success Rate (Grasp SR): The percentage of successful grasps during the task execution.

    \item     Lifting Success Rate (Lift SR): The proportion of episodes where the object is successfully lifted after grasping.

    \item     Re-orientation Success Rate (Orient SR): The proportion of episodes where the object is successfully rotated or adjusted into the target pose (e.g., aligning the hammer for use).
\end{itemize}

We also evaluate training time in experiments, as detailed in Appendix~\ref{subsection:results_training_time}.

\begin{table}[htbp]
    \centering
    \small  
    \resizebox{0.48\textwidth}{!}{  
    \begin{tabular}{@{}l*{2}{c}|*{2}{c}@{}}  
        \toprule
        & \multicolumn{2}{c|}{Without Affordance} & \multicolumn{2}{c}{With Affordance} \\
        \cmidrule(r){2-3} \cmidrule(l){4-5}
        \textbf{Environment} & \textbf{PPO} & \textbf{SAC} & 
        \textbf{PPO} & \textbf{SAC} \\
        \midrule
        Task 1 & $55.3 \pm 1.8$ & $51.8 \pm 1.3$ & $\textbf{73.2} \pm 1.1$ & $68.3 \pm 0.5$  \\
        Task 2 & $37.9 \pm 0.5$ & $28.3 \pm 1.1$  & $\textbf{65.6} \pm 1.1$ & $57.9 \pm 0.7$ \\
        Task 3 & $68.6 \pm 0.5$ & $75.0 \pm 1.2$ & $69.3 \pm 0.6$ & $\textbf{75.1} \pm 1.1$  \\
        \bottomrule
    \end{tabular}
    }  
    \caption{Overall success rate (percentage) for different tasks.}
    \label{tab:success_rate}
\end{table}

\subsection{Quantitative Results and Analysis}
\subsubsection{Overall Success Rate}
As shown in Tab.~\ref{tab:success_rate}, incorporating object affordance information consistently improves success rates across all tasks and policy types. On average, the use of object affordance leads to a 15.4\% absolute improvement in task success rate compared to training without object affordance guidance.

The benefit of affordance integration is particularly significant in Task 2, where the success rate for PPO increases from 37.9\% to 65.6\%, and for SAC from 28.3\% to 57.9\%. This indicates that affordance guidance is especially critical in tasks that require semantic understanding of object parts, such as grasping objects by their handles or functional regions. In contrast, for more dynamic or pose-based tasks like Task 3 (e.g., hammer re-orientation), the relative gains are smaller, as the policies may rely more on motion cues than object-part semantics.

These results highlight the importance of task-relevant affordance signals in improving policy learning and generalization in complex manipulation scenarios.

\subsubsection{Success Rates and Reward Training Curves}
Fig.~\ref{fig:All_Policy_Comparison} presents performance curves for each configuration on the Cube Grasping and Lift, Functional Grasping and Lift of Jug, and Hammer Grasping and Re-orientation tasks. Across all tasks, policies augmented with affordance information exhibit faster convergence and higher final success rates.

Sub-task analysis reveals that object affordance inputs consistently improve early-stage grasp stability (Grasp SR) and downstream manipulation stages (Lift SR and Orient SR).

\subsubsection{Summary of Key Findings}

Based on the results presented above, we identify and analyze three central questions that highlight the interaction between object affordance guidance, task structure, and RL algorithm performance:

\textbf{When does affordance guidance provide the most benefit?}~Affordance guidance is most effective in tasks that require selecting semantically meaningful contact points, such as grasping objects by their handles or flat surfaces. In these tasks, both PPO and SAC perform relatively poorly without affordance cues due to the high dimensionality of the contact space. Our experiments show that introducing affordance information leads to substantial gains in grasping and lifting success rates (e.g., a 27.7\% absolute improvement in jug grasping for PPO), highlighting the importance of affordance in reducing semantic ambiguity during action selection.

\textbf{How does affordance guidance interact with exploration in reinforcement learning?}~Affordance information acts as a semantic prior that constrains exploration to functionally relevant regions of the object. This is especially important in tasks where a large portion of the state-action space corresponds to ineffective interactions. Our results demonstrate that without affordance, exploration is inefficient and policies converge slowly. With affordance, agents focus exploration in task-relevant areas (e.g., near handles), substantially improving sample efficiency and final performance.

\textbf{Are certain RL algorithms more suited for affordance-guided tasks?}~Our experiments reveal a complementary pattern: PPO benefits more from affordance in structured tasks (e.g., lifting), while SAC performs better in tasks that inherently require extensive exploration, such as object re-orientation. This indicates that affordance guidance compensates for the exploration limitations of PPO in complex semantic tasks, whereas SAC's entropy-driven policy already supports broad exploration but still gains modest benefits from affordance-based constraints.

\subsection{Limitations}
We also evaluated our method on more complex tasks, such as opening a microwave door and pick-and-place cube into microwave using dual-arm robot. However, we observed that mesh penetration 
during training had a significant impact on performance. The learned policy tended to exploit these unrealistic penetrations—passing the robotic hand through the microwave door—rather than performing stable manipulation using the designated affordance regions. Addressing this issue is left for future work. More details are summarized in Appendix~\ref{subsection:mesh_penetration}.

\section{Conclusions and Future Work}
\label{sec:conculsion}

In this work, we present a reinforcement learning framework for dexterous manipulation that explicitly incorporates object affordance information to guide both functional grasp pose generation and policy optimization. By integrating functional grasp candidates into the RL pipeline and formulating an object affordance-aware reward structure, our method effectively constrains exploration to semantically meaningful regions of the object. Experimental results across a range of manipulation tasks demonstrate that object affordance guidance substantially improves success rates and RL convergence speed, particularly in semantically complex tasks such as jug and hammer manipulation. On average, the use of object affordance leads to a 15.4\% absolute improvement in task success rate compared to training without object affordance guidance. Moreover, our analysis reveals complementary strengths of PPO and SAC under object affordance guidance, offering insights into the interplay between structured priors and policy exploration. While our approach exhibits robust performance across varied settings, challenges such as physical plausibility in mesh-heavy environments remain, motivating future extensions toward more realistic simulation and contact modeling.

%

\bibliographystyle{IEEEtran}
\bibliography{main}

\appendix
\subsection{Results of Training Time}
\label{subsection:results_training_time}
Table~\ref{tab:headless_training_time} reports the average headless training time (in minutes) for PPO and SAC across all three tasks, both with and without affordance guidance.

We observe that incorporating affordance information consistently reduces training time across all settings. On average, PPO with affordance is approximately 4.6\% faster than its baseline, while SAC achieves a similar reduction of about 6.8\%. The most notable reduction is seen in Task 3 (e.g., hammer re-orientation), where affordance-guided SAC reduces training time from 178.9 to 166.2 minutes.

\begin{table}[htbp]
    \centering
    \small  
    \resizebox{0.48\textwidth}{!}{  
    \begin{tabular}{@{}l*{2}{c}|*{2}{c}@{}}  
        \toprule
        & \multicolumn{2}{c|}{Without Affordance} & \multicolumn{2}{c}{With Affordance} \\
        \cmidrule(r){2-3} \cmidrule(l){4-5}
        \textbf{Environment} & \textbf{PPO} & \textbf{SAC} & 
        \textbf{PPO} & \textbf{SAC} \\
        \midrule
        Task 1 & 119.3 & 141.5 & 115.8 & 136.8\\
        Task 2 & 122.3 & 145.2 & 118.7 & 137.8  \\
        Task 3 & 156.7 & 178.9 & 142.5 & 166.2 \\
        \bottomrule
    \end{tabular}
    }
    \caption{Average headless training time (in minutes) for different algorithms across various environments}
    \label{tab:headless_training_time}
\end{table}
\subsection{Failures because of mesh penetration}
\label{subsection:mesh_penetration}

We also trained a more complex task using multi-agent reinforcement learning, in which a dual-arm robot is required to open the microwave door and place a small cube inside. The success rate of this task is shown in Tab.~\ref{tab:success_rate_microwave}. Notably, the performance is noticeably affected by mesh penetration during the interaction between the robot model and the microwave. Such interpenetrations often lead to unrealistic behaviors and reduce the reliability of the learned policy.

\begin{table}[!htbp]
    \centering
    \small  
    \resizebox{0.48\textwidth}{!}{  
    \begin{tabular}{@{}l*{2}{c}|*{2}{c}@{}}  
        \toprule
        & \multicolumn{2}{c|}{Without Affordance} & \multicolumn{2}{c}{With Affordance} \\
        \cmidrule(r){2-3} \cmidrule(l){4-5}
        \textbf{Environment} & \textbf{HAPPO} & \textbf{MAPPO} & 
        \textbf{HAPPO} & \textbf{MAPPO} \\
        \midrule
        Task 4 & $52.7 \pm 0.9$ & $23.2 \pm 0.7$ & $68.1 \pm 1.0$ & $55.0 \pm 0.3$  \\
        \bottomrule
    \end{tabular}
    }  
    \caption{Success rate (percentage) for opening microwave and place cube into the microwave.}
    \label{tab:success_rate_microwave}
\end{table}

\begin{figure}[!htbp]
    \centering
    \includegraphics[width=0.6\linewidth]{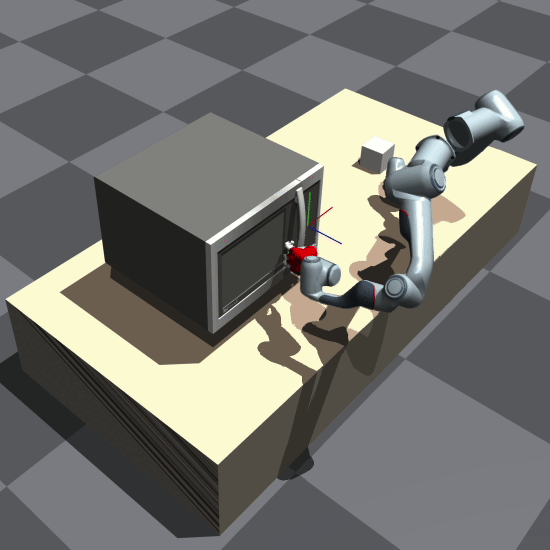}
    \caption{Scene of task 4 for opening microwave and picking, placing cube.}
    \label{fig.failures_microware_open}
\end{figure}

Although the use of object affordance guidance also improves task success rates, as shown in Table IV, we observed that the robot often exploits mesh penetration during manipulation to complete the task. This undermines the physical realism of the learned behavior. In future work, we plan to incorporate mesh-penetration-aware reward functions to discourage such physically implausible interactions and promote more stable and realistic manipulation strategies.

\end{document}